\documentclass[conference]{IEEEtran}

\usepackage{array}
\usepackage{epsfig}
\usepackage{graphicx}
\usepackage{booktabs}
\usepackage{subfigure}

\ifCLASSINFOpdf
\else
\fi
\begin{document}
%
\title{A Data Mining framework to model Consumer Indebtedness with Psychological Factors}

\author{\IEEEauthorblockN{Alexandros Ladas}
\IEEEauthorblockA{School of Computer Science\\
University of Nottingham\\
Nottingham, UK, NG8 1BB\\
psxal2@nottingham.ac.uk}
\and
\IEEEauthorblockN{Eamonn Ferguson}
\IEEEauthorblockA{School of Psychology\\
University of Nottingham\\
Nottingham, UK,NG7 2RD\\
eamonn.ferguson@nottingham.ac.uk}
\and
\IEEEauthorblockN{Uwe Aickelin and Jon Garibaldi}
\IEEEauthorblockA{School of Computer Science\\
University of Nottingham\\
Nottingham, UK, NG8 1BB\\
\{uwe.aickelin,jon.garibaldi\}@nottingham.ac.uk
}
}


%


\maketitle

\begin{abstract}
Modelling Consumer Indebtedness has proven to be a problem of complex nature. In this work we utilise Data Mining techniques and methods to explore the multifaceted aspect of Consumer Indebtedness by examining the contribution of Psychological Factors, like Impulsivity to the analysis of Consumer Debt. Our results confirm the beneficial impact of Psychological Factors in modelling Consumer Indebtedness and suggest a new approach in analysing Consumer Debt, that would take into consideration more Psychological characteristics of consumers and adopt techniques and practices from Data Mining.
\end{abstract}


%
\IEEEpeerreviewmaketitle

\section{Introduction}
As Consumer Debt has risen to become a significant problem of our modern societies, especially in developed countries, it triggered the interest of Academic Community, which tried to provide reasonable explanations for the ``nature'' of this complex problem. In Economics, modelling Consumer Indebtedness traditionally adopted the ``rational'' behaviour of debtors \cite{ottaviani2011impulsivity}, limiting the research on socio-economic variables. But recent interdisciplinary studies show  that the behaviour of consumers deviates from the ``rational'' model \cite{ottaviani2011impulsivity} and that level of debt prediction is not a function of economic factors exclusively \cite{webley2001life,lea1995psychological,wang2011demographics}.

Now Consumer Indebtedness is considered a phenomenon with distinct facets \cite{kamleitner2012credit,stone2006indicators}, which is influenced by several psychological factors and entails a magnitude of privetal,  societal and economic implications. The incorporation of disciplines from Psychology in Consumer Debt Analysis revealed the significance of psychological factors in modelling Consumer Indebtedness, by associating a series of Personality traits, attitudes, beliefs and behaviours to Consumer Debt.

However, as revolutionary the findings from these studies are, the research in this field suffers from certain limitations. First of all, as the authors in \cite{livingstone1992predicting} point out that no clear and conceptual model of Consumer Indebtedness has yet emerged despite the fact that many factors influencing Consumer Debt have been proposed in literature, which so far has focused on a few or a subset of demographic, economic, psychological and situational factors, thus limiting, its predictive ability and its generalizability \cite{stone2006indicators}. This may have been due to the reason that no database exists in literature that includes an extensive list of these factors as well as sufficient information to define dependent variables representing financial indebtedness. Secondly, the research has been carried out by statistical models with weak predictive ability \cite{ladas2014augmented} that face numerous difficulties when they deal with real world data. This raises questions regarding the reliability of the findings in the literature. Finally, most of the research has been conducted on a limited number of observations making hard to consider the findings as representative.

As the incorporation of disciplines from the rich theory of Psychology proved to be successful for the purposes of Consumer Debt Analysis, we believe that the latter has a lot to gain from embracing techniques from Data Mining.  The careful data preprocessing, the powerful models and reliable evaluation techniques, comprise a complete and sophisticated toolbox to analyse complex real world data, like socio-economic data, and can guarantee representative and meaningful Knowledge discovery.

Towards this direction, in this work we examine the multifaceted aspect of Consumer Indebtedness by exploring the impact of Psychological Factors upon models that are based traditionally on Demographic and Economic Data, within a complete Data Mining framework. We apply our methods and techniques on  a very detailed dataset, DebtTrack, which is a socio-economic online survey conducted by YouGov and also includes items that measure Psychological Factors, like Impulsivity and Risk Aversion. This dataset gives us the opportunity to research the multifaceted nature of Consumer Indebtedness as it is described in \cite{kamleitner2012credit,stone2006indicators}. In more detail, we pre-process the data to reduce dimensionality and remove the noisy data as well as to extract and verify the Psychological Factors of the data. Afterwards we utilise three Data Mining models from different families of predictive modelling, namely Logistic Regression, Random Forests, and Neural Networks to assess the contribution of Psychological Factors in Consumer Debt Analysis in an extensive number of experiments. The superior performance of these models in modelling Consumer Indebtedness has been established in our previous study \cite{ladas2014augmented}and therefore they pose as reliable models to draw conclusions from safely.

Our results show that Psychological Factors are significant predictors of Consumer Indebtedness confirming the multifaceted nature of the problem \cite{kamleitner2012credit,stone2006indicators}. Our final model manages to exhibit a good predictive ability and thus can be considered as the first step towards a complete and well specified model of Financial Indebtedness, as described in \cite{stone2006indicators} which should include all the additional and necessary factors for Consumer Debt Analysis. Despite the fact that the data transformations were not able to improve the performance of the models as has happened in \cite{ladas2014augmented}, in the case of Demographics the transformed dimensions managed to capture the information of the attributes with their performance being comparably close to the one of Demograpic variables and therefore can pose as good dimensionality reduction. Overall our Data Mining framework provides an elaborate toolbox to analyse Consumer Debt and successfully draw safe conclusion that can contribute to the research.
 
The rest of the paper is organised as following. In 2nd section we discuss the related work on modelling Consumer Indebtedness and in the 3rd section we present the DebtTrack dataset and the variables of interest. The 4th section analyses the basic pre-processing steps we did to handle noisy data and the extraction and validation of the Psychological Factors whereas the in the 5th we proceed with experimental validation of the Psychological Factors. Finally in the 6th section we conclude our work.

\section{Related Work}
The research for Consumer Debt Analysis has been mainly focused on answering three fundamental questions as they are formulated in \cite{wang2011demographics}:
\begin{enumerate}
\item Which factors discriminate debtors from non-debtors?
\item Which factors affect how deep consumers go into debt?
\item Which factors influence the repayment of debt?
\end{enumerate}
Answering these questions has led to the discovery of a series of factors that are associated with Consumer Indebtedness. Among them, we can identify personality traits like Self Control or Impulsivity, other psychological factors like Risk Aversion and attitudinal data regarding debt, credit use and money. The extensive list of factors that is presented in the literature of this field comes to supplement well researched socio-economic factors that have been traditionally used in economic models to explain Consumer Debt like, work status, net wealth and number of children in household \cite{ottaviani2011impulsivity} and income, age, gender, education in \cite{kamleitner2012credit}.  While an abundance of factors has been suggested to be related to Consumer Debt, articles like \cite{wang2011demographics, kamleitner2012credit, stone2006indicators, kamleitner2007consumer} provide a nice review of the factors proposed and nice overview of the ongoing research in the field.

Considering Impulsivity or Self Control the work in \cite{gathergood2012self} revealed a significant association with Consumer Debt, especially in the cases of credit cards, mail order catalogues, home credit and payday loans. Apart from the significance of Impulsive spending, in the same work it has been linked to Impulsive Personality that led to income shocks. In \cite{ottaviani2011impulsivity} Impulsivity was found a strong predictor of Unsecured Debt but not Secured Debt, like mortgages and car loans. The rationale behind this, stems from the fact that Secured Debt affects decisions that last for a long time and therefore is linked to Life-Cycle theory \cite{webley2001life}, according to which consumer enter into debt on rational grounds in order to maximise utility and thus is not associated with Impulsive behaviour which favours short run benefits like the ones Unsecured Debt usually provides.
 
Finally in \cite{stone2006indicators} a multifaceted model of Consumer Indebtedness is confirmed where all the groups of variables are important in a combined model and not as single predictors. Among the predictors identified, age, sex and ethnicity for the group of Demographics and the income and financial activities for the group of Financial variables found to accompany a series of Psychological and Sociological factors.

The problem of all this work summarised above is that it is characterised by a weak statistical modelling with the proportion of variance being explained in most of the models being around 10\%. Based on this weak predictive modelling and the fact that the models are built on a limited number of observations, we are unsure whether to regard these findings as reliable since the suggested models fail to explain the variance that exists in the data and the small number of instances
cannot be considered representative enough. 

On the other hand the necessity of incorporating Data Mining in the analysis of socio-economic data is emphasised in \cite{helbing2011social}. Since Data Mining offers models with strong predictive power, their utilisation can provide better answers on the study of complex systems where linear, conventional and straight forward modelling is usually inappropriate.Data Mining . In \cite{ladas2014augmented} Random Forests and Neural Networks exhibited superior performance in modelling Consumer Indebtedness than the ones presented in Economic Psychology literature. Random Forests have been utilised in marketing applications like \cite{ghose2011estimating} where a model measuring the impact of the reviews of products in sales and perceived usefulness was constructed, whereas Neural Networks have been used in Economics for for stock performance modelling \cite{nicholas1994stock} and for credit risk assessment \cite{atiya2001bankruptcy}. This highlights the contribution of Data Mining in developing accurate quantitative prediction models for Economic application, as dictated in \cite{atiya2001bankruptcy}. 
 
 \begin{figure*}[!t]
\centering
\includegraphics[scale=0.5]{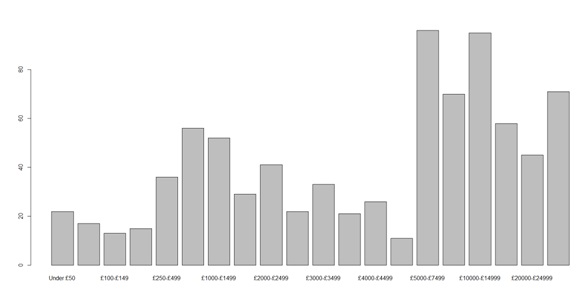}
\caption{Histogram of Unsecured Debt}
\label{fig 1}
\end{figure*}

\section{DebtTrack Survey}
\subsection{General Overview}
DebtTrack Survey \cite{gathergood2012self} is a quarterly repeated cross-section survey of a representative sample of UK households. It is been carried out by the market research company YouGov and it focuses on the consumer-credit information of households. It is consisted of 2084 households and in its 85 questions it covers  household demographics, labour market information, income and balance
sheet details. The consumer credit data are particularly detailed providing information about the
number and type of consumer credit products , outstanding balances for each debt product , monthly payments and whether they are one month or three months in arrears on the product as well as
the value of arrears. The most interesting aspect of this dataset is that it contains 28 questions that measure Psychological characteristics like Impulsivity and Risk aversion, which gives us the opportunity to explore the multifaceted ``nature'' of Consumer Indebtedness on a complete dataset by the standards specified in \cite{stone2006indicators}

The variables of interest from Demographics and Financial attributes can be seen in Table 1. The survey contains a much bigger number of Demographic and Financial questions but they had a big number of uncertain answers that is why they were excluded from our analysis. By Uncertain answers we refer to answers that were given to questions, like ``I don't know'' or ``Prefer not to answer'', which while they cannot be regarded as missing values they definitely do not provide any valuable information. In the variables included in Table 1 we can see that some of the Financial attributes exhibit a big number of uncertain answers that can be regarded as noisy data. It is worth noted that the \emph{Guardian} and \emph{Insurance} variables are actually groups of smaller Boolean variables, five for the case of \emph{Guardian} and 11 for the \emph{Insurance}.

\begin{table}[btp]
  \centering
  \caption{Demographic and Financial attributes of DebtTrack}
    \begin{tabular}{ll}
    \textbf{Attribute} & \textbf{Unknown Answers} \\
	\hline    
    \multicolumn{2}{c}{Demographic} \\
    Marital\_Status                  & 17 \\
     Emp\_Status                       & 32 \\
    Age\_Sex                         &  \\
    Social\_Grade                    &  \\
    Education                       & 32 \\
    Guardian &  \\
    \multicolumn{2}{c}{Financial} \\
    Household Income                  & 492 \\
     Income                            & 548 \\
    Liquid Assets                      & 340 \\
    House\_Status                    & 30 \\
    Insurance &  \\
    \hline
    \end{tabular}%
  \label{tab:addlabel}%
\end{table}%
 
Consumer Indebtedness can be modelled in very different ways, as it can be seen throughout the literature. In this work we are going to use the Unsecured Debt as dependent variable of the models. The choice was based on the well established relationship between Unsecured Debt and Impulsivity \cite{ottaviani2011impulsivity}, which is a Psychological factor that is verified in our dataset. Unsecured debt is a categorical variable with 21 levels. As we can see in Fig 1. the class variable is heavily imbalanced and most of the classes have very few members so they will be under-represented. For this reason we decided to move to a two-class classification (In Debt, No Debt) which is more balanced with 652 consumers being in Debt and 601 build models that can discriminate the debtors from non-debtors, a fundamental research question of Consumer Debt Analysis \cite{wang2011demographics}. As the level of debt prediction is another important direction of this research we also attempt to build a three-class classification model (No Debt, High, Low)for predicting, but that is also imbalanced. No Debt has 600 member whereas the other two class around 300. Imbalanced classes is a traditional problem in Data Mining \cite{chawla2005data,mollineda2007class} but while a lot of sophisticated methods have been presented in literature they are usually not suitable for a multiclass model \cite{mollineda2007class}. For this reason we decided to use a simple under-sampling of the ``No Debt'' class and we chose 300 instances randomly to be the new ``No Debt''. Despite its simplicity under-sampling is considered a very reliable method \cite{chawla2005data,mollineda2007class}.  The new sample is representative of the whole ``No Debt'' class as statistical tests showed, so there is no loss of information.
  
\begin{figure*}[!t]
\begin{center}
	\mbox{
	\subfigure[Demographic Categories]
	{\epsfig{file=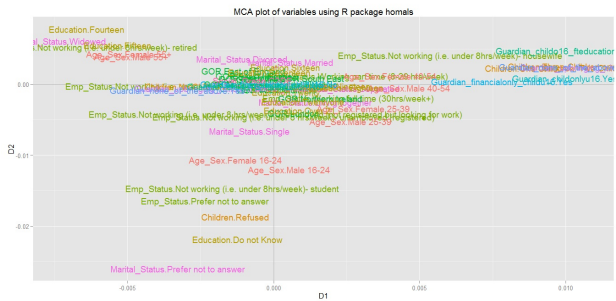,scale=0.23}}		
	\subfigure[Financial Categories]
	{\epsfig{file=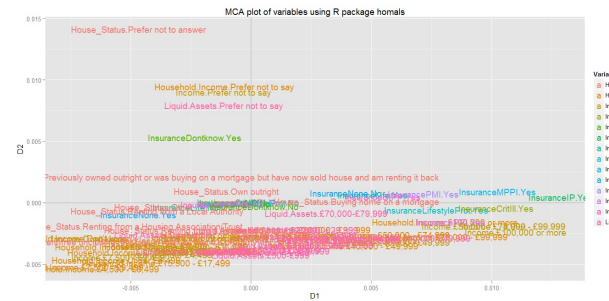,scale=0.23}}	}
\caption{Category plots of DebtTrack variables in the 2-dimensional space created by Homogeneity Analysis}
\end{center}
\end{figure*}

\begin{figure*}[!t]
\begin{center}
	\mbox{
	\subfigure[Demographic Categories]
	{\epsfig{file=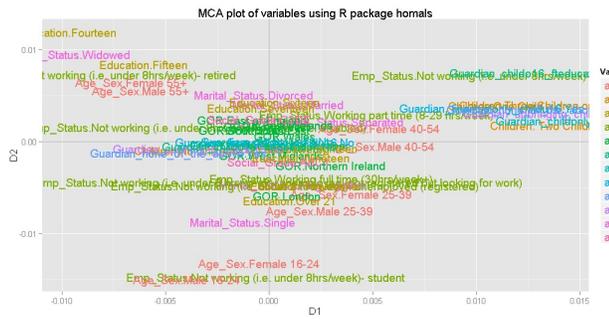,scale=0.3}}		
	\subfigure[Financial Categories]
	{\epsfig{file=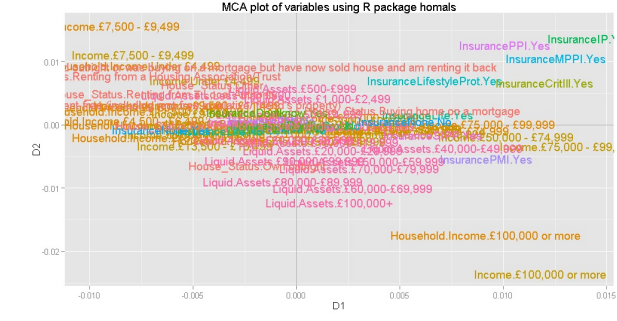,scale=0.3}}	}
\caption{Category plots of DebtTrack variables in the 2-dimensional space created by Homogeneity Analysis on the reduced dataset}
\end{center}
\end{figure*}

\section{Data Pre-processing} 
\subsection{Handling Noise}
As we were unsure whether to include in our models the instances that contained a large number of uncertain answers we performed a Homogeneity Analysis on the Demographic and Financial attributes in order to get a better insight of the data. Homogeneity analysis (Homals) \cite{de2009gifi,de2007homogeneity}, is a method  very similar to correspondence analysis. Homals maps a set of categorical variables into a number of p dimensions. In this p-dimensional space similar, the category points serve as centers of gravity of the objects that share the same category. That guarantees that similar objects will be placed close to each other in this representation. It achieves that, based on the criterion of minimizing the departure from homogeneity. This departure is measured by the sum of squares of the distances between the object scores and their corresponding categories. Thus the problem is reduced to minimise these distances in the p-dimensional space. Certain constraints guarantee that the representation will be centered and that the object scores will be orthogonal.

A view of of the 2-dimensional space created by Homals can be seen in Fig 2. and reveals associations between categorical levels. We can notice that the categorical levels of ``Don't know'' and `` Prefer not to answer'' form an outlier cluster away from the other categories which seem to be centered. In the case of Demographics the cluster is positioned below the center and in case of Financial attributes above the center. That means that the people who preferred not to reveal any information, they did so, systematically to a series of questions. So these instances cannot contribute any valuable information to the modelling and it was decided to be removed from the further analysis moving to a reduced data size.

After the removal of these noisy instances the data size reduced to 1253 instances from the 2084 it originally had. Chi-square tests were utilised to test the differences between the expression of the same categorical variables in the sample and in the original population. All the results showed that the reduced dataset hold the same information with the original dataset as no big differences in the distribution of categorical levels were noticed, with the bigger one being 5\%. That is a very small proportion of change considering that almost half of the dataset was removed. The category plot of the smaller dataset can be seen in Fig.3. Removing these instances caused both category plots to be more dispersed giving more discriminative power to the dimensions of Homogeneity Analysis. This way the representation can differentiate easier among categories by assigning a more meaningful score to the representation. The biggest impact of our strategy can be seen in the biplot of the Financial attributes where the small and dense cluster in the bottom of the plot in Fig. 1 has changed to a bigger cluster centered in the plot that occupies most of the space. In demographics we can see that the deletion of these instances caused the biplot to be more understandable since it seems that a formation of clusters starts to appear. This cannot be noticed in the case of Financial dimension where the categories of Financial Attributes appear to be very close to each other.

As this representation of Homals proved to be successful in our previous work on modelling Consumer Indebtedness \cite{ladas2014augmented} as they increased the predictive power of all our models we decided to keep these transformed variables, getting two Demographic Dimensions and two Financial Dimensions.



\subsection{Factor Analysis on Psychological Items}

In an effort to understand the structure of the Psychological items of the survey and to validate the theoretical constructs they were trying to assess we performed an Exploratory Factor Analysis on them (EFA). The 28 Psychological Items in the survey cover psychological aspects like Self Control or Impulsivity, Risk Management/Knowledge and Risk Aversion. Since the number of theoretical constructs was unclear in the original data, we chose EFA instead of CFA in order to find of latent structure that can represent the 28 items. After the factors have been specified we can verify which of the theoretical constructs are actually being assessed in this survey.

In order to define the optimal number of factors that describe best the Psychological items, we utilised the Scree test and Parallel Analysis two widely used techniques in applied EFA for determining the number of latent factors \cite{fabrigar1999evaluating}. In the Scree test the eigenvalues of the correlation matrix of the variables are plotted in descending order. The number of optimal factors is then chosen by the number of eigenvalues that precede the last substantial drop in the graph. In Parallel Analysis the eigenvalues of the same correlation matrix are compared to eigenvalues of the correlation matrices of randomly generated datasets with same size as the original. The number of factors is chosen by the number of eigenvalues that is bigger than the number of eigenvalues of random data. In Fig. 4 you can see the results of the Scree Test and Parallel Analysis suggesting that five should be the optimal number of latent factors.

\begin{figure}[!t]
\centering
\includegraphics[scale=0.5]{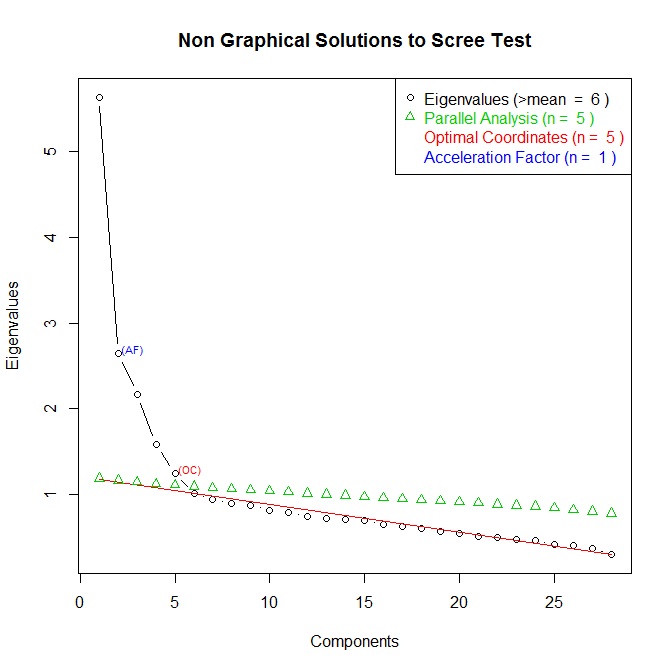}
\caption{Scree plot of the eigenvalues of the Psychological Items and random data}
\label{fig 4}
\end{figure}

So we proceeded with a 5-factor model to represent the Psychological items. The factor loadings of the five factors on the Psychological items  can be seen in Table 2. We chose an orthogonal rotation of the factors despite the many advantages of obliques rotation \cite{fabrigar1999evaluating} because uncorrelated predictors is much desired in Data Mining models. The 5-factor model explains 36\% of the total variance and the first factor explains almost the half of this amount.

\begin{table*}[htbp]
  \centering
  \caption{Factor loadings of 5-factor model on Psychological Items}
    \begin{tabular}{llllll}
    \toprule
          & \textbf{Factor 1} & \textbf{Factor 2} & \textbf{Factor 3} & \textbf{Factor 4} & \textbf{Factor5 } \\
    \midrule
    Q70r1 & 0.75  &       &       &       & -0.195 \\
    Q70r2 & 0.689 &       &       &       & -0.123 \\
    Q70r3 & 0.732 &       & -0.112 &       &  \\
    Q70r4 & -0.168 &       & 0.201 &       & 0.391 \\
    Q70r5 & 0.681 &       &       &       & -0.426 \\
    Q70r6 & 0.583 &       &       &       & -0.179 \\
    Q70r7 & -0.428 & 0.133 &       &       & 0.656 \\
    Q70r8 & 0.522 & -0.138 &       & 0.128 &  \\
    Q70r9 & 0.646 &       & -0.209 & 0.123 &  \\
    Q70r10 &       &       & 0.275 & 0.412 &  \\
    Q70r11 & -0.154 &       & 0.592 & 0.156 &  \\
    Q70r12 &       & -0.102 & 0.408 & 0.334 &  \\
    Q70r13 & 0.11  &       & 0.305 & 0.215 &  \\
    Q70r14 & 0.107 &       &       & 0.648 &  \\
    Q70r15 &       &       &       & 0.592 &  \\
    Q70r16 &       &       & 0.428 &       &  \\
    Q70r17 & -0.183 & 0.118 & 0.46  & -0.118 &  \\
    Q71r1 & 0.194 & -0.389 & 0.165 & 0.129 & 0.174 \\
    Q71r2 &       & 0.601 &       &       &  \\
    Q71r3 &       & 0.643 &       &       & 0.126 \\
    Q71r4 & -0.383 & 0.425 &       &       & 0.328 \\
    Q71r5 & 0.326 &       &       &       &  \\
    Q71r6 & -0.129 & 0.209 & 0.414 &       &  \\
    Q71r7 &       & 0.373 & 0.109 &       & -0.161 \\
    Q71r8 & 0.636 & -0.154 &       &       & -0.107 \\
    Q71r9 & -0.409 & 0.479 & 0.222 &       & 0.121 \\
    Q71r10 &       & -0.122 & 0.395 & 0.104 & 0.205 \\
    Q71r11 & 0.428 & 0.146 &       &       & -0.113 \\
    \bottomrule
    \end{tabular}%
  \label{tab 2}%
\end{table*}%

After looking at how the Psychological Items are grouped together based on the 5-factor model we can see that the first factor loads significantly on items measuring Impulsivity and Self Control, the 2nd factor on Risk Aversion items, the 3rd on Organisational Responsibility the 4th on Risk Management Belief and finally the 5th one on Planful Saving. The renamed factors and their Cronbach's alpha can be seen in Table 3. Cronbach's alpha measures the internal consistency of each factor, meaning how intercorrelated are the items that loads on. In other words, it tests if items loaded on a factor measure the same construct the latent factor represents. Cronbach's alpha takes value from 0 to 1 with a taking values above 0.7 being considered good, between 0.6 and 0.7 as acceptable and between 0.5 and 0.6 as poor. In similar fashion, as we can see  in Table 3 only Impulsivity can be considered a reliable measure. Risk aversion and Organisational Responsibility can be regarded as acceptable and the last two as poor.

\begin{table}[htbp]
  \centering
  \caption{Identified Psychological Factors and their Internal Consistency}
    \begin{tabular}{ll}
    \toprule
    \textbf{Factors} & \textbf{Cronbach's alpha} \\
    \midrule
    Impulsivity & 0.86 \\
    Risk Aversion & 0.64 \\
    Organisational Responsibility & 0.61 \\
    Risk Management Belief & 0.57 \\
    Planful Saving & 0.57 \\
    \bottomrule
    \end{tabular}%
  \label{tab 3}%
\end{table}%

 \section{Experimental Evaluation of Psychological Factors}
 \subsection{Experimental Setup}
After having identified the Psychological Factors in the dataset we can now evaluate their contribution in modelling Consumer Indebtedness. For this reason we group the variables into three groups, Demographics, Financial and Psychological Factors. For the Demographics and Financial groups we also have obtained their transformed representations from the Homogeneity Analysis whereas the Psychological Factors include the five factors identified by Exploratory Factor Analysis. Then we check each of the five groups of variables individually in order to understand their predictive power. Then we proceed in a stepwise fashion starting from Financial variables (Step 1) and we continue adding Demographics (Step 2) and Psychological factors (Step 3) in the process, to examine carefully the accuracy of the multifaceted model as this develops gradually. This is repeated for the transformed variables. Finally we compare the differences in the performance of the models between Step 3 (Financial and Demographics and Psychological) and Step 2 (Financial and Demographics only) in order assess the impact of Psychological factors on modelling Consumer Indebtedness. Since in Step 2 the models are based on socio-economic variables only and in Step 3 Psychological factors are also added, this comparison points out the importance of including Psychological Factors in the traditional Economic modelling of Consumer Indebtedness. The significance of the impact is measured by statistical significant testing.

As classifiers, three different Data Mining models with different characteristics are used, Multinomial Logistic Regression, Random Forests and Neural Networks. Multinomial Logistic Regression belongs to the family of linear Classifiers and measures the relationship between a categorical dependent variable and one or more independent variables, by using probability scores as the predicted values of the dependent variable. Random Forests is an example of ensemble learning that
generates a large number of Decision Trees, built on different samples with bootstrap methods that allow re-sampling of instances, and aggregate the results. The difference from being an ensemble of Decision Trees is that when a split on a node is to
be decided, a specific number of the attributes chosen randomly can participate
as candidates and not all of them. The 3rd model is Neural Networks which is considered a non-linear classifier. Neural Networks connect the input variables (predictors) to the output variable through a network of neurons organised in layers, where each neuron of every layer is fully connected to the output of all the neurons in the previous layer. The output of each neuron is a non-linear transformation of the sum of all its inputs. The three different classifiers can test different aspects of modelling Consumer Indebtedness and reveal important characteristics of this dataset.

Finally for evaluating the performance of all the models we build and estimating their accuracy, we use a repeated 10-fold cross validation method, which is widely used evaluation method in Data Mining. 10-fold cross validation gives a more generalised and reliable evaluation of the model as it doesn't allow over-fitting, which means that the model will exhibit the same performance on unseen data. The repeated version of 10-fold cross validation gives an even more reliable estimate of performance of the model as it is not affected by the partitioning of the folds. 

All of our models are built in R using the caret package \cite{kuhn2013applied} and for Neural Networks we use one hidden layer of neurons and we tune the number of neurons in this layer by keeping the model with the best performance from all the possible models with neurons varying from one to ten. Also ten is the number of repetitions for 10-cross fold validation.

\subsection{Two-class models}

In Fig. 5 we can see the performance of the models that are built on a single group of variables. We can see that Financial variables hold the strongest predictive ability among the five different groups, while Psychological factors come 2nd and Demographic Variables 3rd. It is clear that Psychological factors hold significant predictive ability and they exhibit better performance than the Demographic variables which were traditionally included in Economic models of Consumer Indebtedness. However, we can also see that the transformed variables show the worst performance of all five groups and especially the biggest drop is manifested in the case of Financial dimensions, a fact that signifies a significant loss of information. In case of Demographics the performance of Demographic dimensions is worse but comparable to the original Demographic variables. Considering the classifier Neural Networks and Multinomial Logistic Regression exhibit similar performance with Random Forests being slightly worse.

 \begin{figure}[!t]
\centering
\includegraphics[scale=0.7]{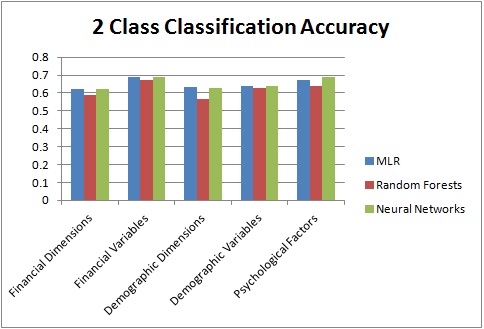}
\caption{Performance of groups of variables in 2-class classification}
\label{fig 5}
\end{figure}

\begin{figure*}[!t]
\centering
\includegraphics[scale=0.5]{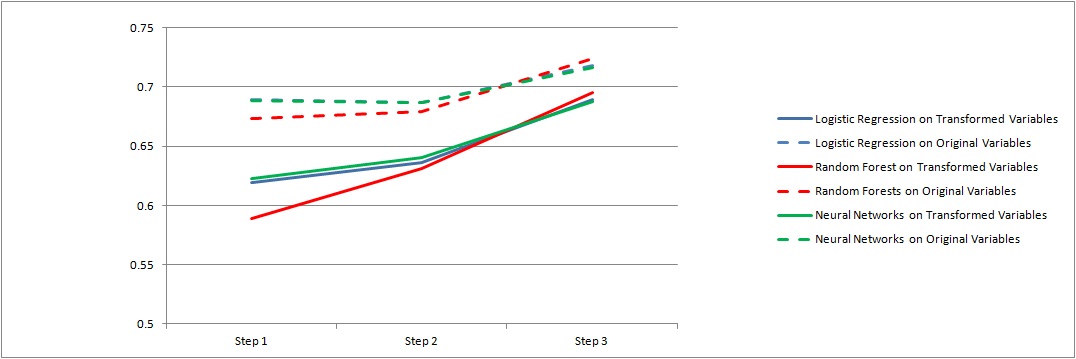}
\caption{Performance of the models in stepwise fashion, Step 1: Financial Variables, Step 2: Demographics Added, Step 3: Psychological Factors Added}
\label{fig 6}
\end{figure*}

The superior performance of the original variables over the transformed ones can be seen more clearly in Fig. 6, where the performance of the model is presented as the groups of variables are added step by step.  All the models exhibit similar behaviour and performance as in each step the performance improves when a group of variables is inserted into the model. That means that every group of variable is beneficial for modelling Consumer Indebtedness. However, the increase in the Accuracy of the model is not very big, around 10\% in almost models except the case of Random Forests for transformed variables where the final model (Step 3) increases its Accuracy around 20\%. This behaviour of Random Forests is evident also in the case of the original variables and interestingly enough, while they exhibit the worst performance in the beginning, as it was also seen in Fig. 5 when examined for groups of variables individually, soon their performance increases in the end they exhibit the best performance at Step 3.

As Psychological Factors seem to increase the performance from all the models improving their predictive ability, their impact was assessed by statistical significant testing. In more detail, all the 100 folds from the ten repetitions of 10-fold cross validation of each model in Step 3 were compared to the corresponding 100 folds of Step 2 with t-tests. The results can be seen in Table 4, where the importance of Psychological Factors in modelling Consumer Indebtedness is pointed out as all the tests were statistically signified with their p-value being much smaller than 0.025. That means that the traditional Economic modelling of Consumer Indebtedness which was exclusively based on socio-economic variables like in Step 2 can benefit from the incorporation of Psychological Factors in Step 3. 
 
\begin{table}[htbp]
  \centering
  \caption{Statistical Significance between Step 2 and 3 for 2-class Classification}
    \begin{tabular}{rr}
    \toprule
    \textbf{Classifier} & \textbf{p value} \\
    \midrule
    \multicolumn{2}{c}{Original} \\
    Multinomial Logistic Regression & $5.153e^{-6}$ \\
    Random Forests & $6.312e^{-11}$ \\
    Neural Networks & $3.918e^{-6}$ \\
    \multicolumn{2}{c}{Transformed} \\
    Multinomial Logistic Regression & $6.558e^{-15}$ \\
    Random Forests & $<2.2e^{-16}$ \\
    Neural Networks & $2.361e^{-15}$ \\
    \bottomrule
    \end{tabular}%
  \label{tab 4}%
\end{table}%

\subsection{Three-class models}

Moving to a three-class classification we can see in Fig 7, that the predictive ability of the variables has dropped around 20\% when compared with the performance of the same variables in the two-class classification. However, the relative predictive ability of the five groups of variables remained the same. Similarly with the two class classification Financial variables appear to be the strongest predictors whereas all the transformed variables continue to exhibit the worst performance of all five groups. The predictive ability of Psychological factors is still better than the the one of Demographic variables but now their difference is much is smaller, while Demographic dimensions achieve again comparable performance with the Demographic variables. Considering the models, Random Forests continue to exhibit the worst performance when they are examined for single group of variables in almost all groups except the Financial Dimensions where Neural Networks are substantially worse.

\begin{figure}[!t]
\centering
\includegraphics[scale=0.7]{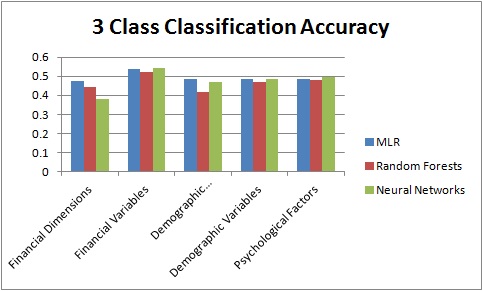}
\caption{Performance of groups of variables in 3-class classification}
\label{fig 7}
\end{figure}

Checking the performance of the models in the stepwise fashion in Fig. 8 verifies the overall worse Accuracy of the three-class classification models comparing to the corresponding two-class classification models. Now the performance of the classifiers is not similar. Random Forests are clearly superior and the Neural Networks exhibit the worse predictive Accuracy. In a similar way as before Random Forests accuracy picks up as more variables are added in the modelling and they are the only classifier that improve the performance in Step 2 in case of original variables. Neural Networks and Multinomial Logistic Regression exhibit a drop in the performance when Demographic variables are added in Step 2. On the other hand the inclusion of Psychological factors is beneficial for all the models, both for original and transformed variables. Now Psychological factors seem to be more important for modelling Consumer Indebtedness than the Demographics. 

\begin{figure*}[!t]
\centering
\includegraphics[scale=0.5]{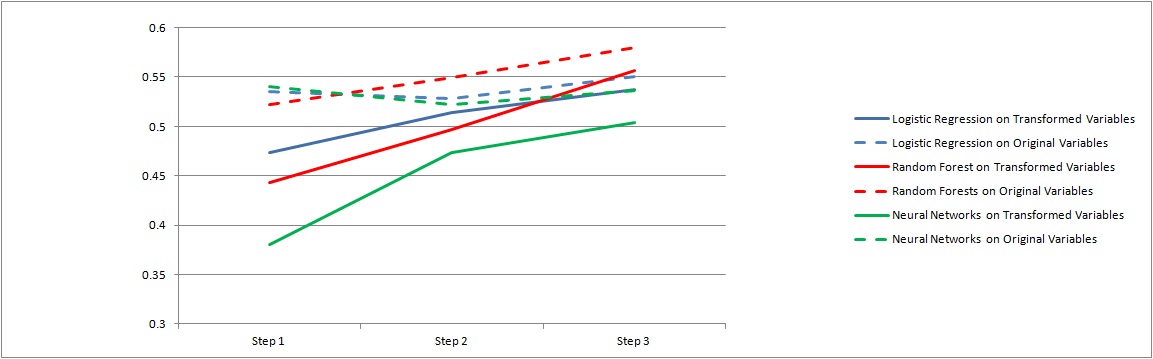}
\caption{Performance of the models in stepwise fashion, Step 1: Financial Variables, Step 2: Demographics Added, Step 3: Psychological Factors Added}
\label{fig 8}
\end{figure*}

Looking at the statistical significance of the increase between Step 2 and 3 in Table 5, where the same tests as in the case of two-class classification were repeated, we see again the very small p-values. Again all the tests were statistically significant showing the importance of Psychological Factors in modelling Consumer Indebtedness.

\begin{table}[htbp]
  \centering
  \caption{Statistical Significance between Step 2 and 3 for 3-class Classification}
    \begin{tabular}{rr}
    \toprule
    \textbf{Classifier} & \textbf{p value} \\
    \midrule
    \multicolumn{2}{c}{Original} \\
    Multinomial Logistic Regression & $0.007919$ \\
    Random Forests & $0.0001202$ \\
    Neural Networks & $9.47e^{-7}$ \\
    \multicolumn{2}{c}{Transformed} \\
    Multinomial Logistic Regression & $0.0005011$ \\
    Random Forests & $8.524e^{-14}$ \\
    Neural Networks & $4.149e^{-6}$ \\
    \bottomrule
    \end{tabular}%
  \label{tab 5}%
\end{table}%

It is now evident from both the 2-class classification and the 3-class classification approaches to model Consumer Indebtedness, that Psychological Factors are important predictors. This has been shown from all the results and statistical tests for all the models for both original and transformed variables. Their contribution to Consumer Debt Analysis seems to be vital and they pose as strong candidates to supplement the existing traditional socio-economic modelling of Consumer Indebtedness. On the other hand the Demographic Variables seem to exhibit worse performance than Psychological Factors not only when they are examined separately but also when they cause a drop in the performance of the models for the cases of Neural Networks and Multinomial Logistic Regression for original variables in the three-class classification, a fact that raises some questions regarding their contribution to the level of Debt prediction. In addition to this, the performance of the models in the two-class classification is superior to the performance of the three-class classification. More accurately the two-class classification achieves a respectable performance from all the models whereas the performance of the models in the three-class classification is below average. While this gives better chances to build models to separate debtors from non-debtors, trying to predict the level of debt remains one of the important questions of Consumer Debt Analysis \cite{wang2011demographics}. Given that the multifaceted ``nature'' of the level of debt prediction was confirmed by showing the importance of the Psychological aspect of the problem, perhaps more research needs to be done in order to produce better models.

Looking at closer at the performance of the classifiers Random Forests manifested the best results especially in the case of three-classification whereas in two-class classification the performance of the classifier is comparable. That means that solving the problem of discriminating debtors from non-debtors does not depend on the characteristics of the three different classifiers. On the contrary, trying to predict the level of Debt seems to benefit from the usage of ensemble learning. 

Finally, the transformations were not able to improve the classifications in any model. That is more clear in the case of Financial dimensions where it seems that the loss of information in this representation is significant. This verifies the absence of discriminative ability of the Financial dimensions as this was noticed on the visual representation of the Financial categories in Fig. 3b. On the other hand the Demographic dimensions, which seem to have more discriminative ability in Fig. 3a, appear as a better representation of the Demographic variables since their performance is comparable to the performance of Demographic variables and that means that they can be used in an effort to reduce dimensionality.

\subsection{Random Forests Analysis}
In our results, it is clear that we can build a good model for discriminating debtors from non-debtors, but building a more reliable and accurate model that predicts the level of Debt requires a deeper and more sophisticated research. From the two-class models Random Forests exhibited the best performance. More accurately the model achieved a 72\% Accuracy, 70\% Sensitivity (true positive rate) and 74\% Specificity (true negative rate). Random Forests also offer a way to assess the Variable Importance by using as measure the mean decrease in Gini Index. Gini index measures the impurity of data, and therefore the attribute that causes the biggest decrease in the impurity is considered a significant predictor. Having a look at the values of this measure which are summarised in descriptive statistics of Table 6, we can notice that more than the 75\% of the attributes have a measured importance that is less than the mean importance of all the attributes. That means that there are few attributes that seem to be very important for the accuracy of the model. 

\begin{table}[htbp]
  \centering
  \caption{Descriptive Statistics of Variable Importance}
    \begin{tabular}{rrrrrr}
    \toprule
    \multicolumn{6}{c}{\textbf{Mean Decrease in Gini Index}} \\
    \midrule
    Min.  & 1st Qu. & Median & Mean  & 3rd Qu. & Max. \\
    0.2091 & 1.406 & 2.479 & 4.179 & 3.338 & 50.83 \\
    \bottomrule
    \end{tabular}%
  \label{tab 6}%
\end{table}%

Taking a closer look in the values of the mean decrease in Gini Index for all the attributes we can spot 
seven attributes that they have much bigger values than the rest of the dataset. These attributes are presented in Table 7. We can see that the values of these attributes in the decrease of Gini Index are many times bigger than the average decrease in Gini index of all the attributes of the dataset. Interestingly enough all the Psychological Factors are included as predictors , especially Impulsivity and Planful Saving. Finding Impulsivity being the strongest predictor among of all the variables confirms the findings of \cite{gathergood2012self,ottaviani2011impulsivity}, where it was mentioned to be strongly associated with Consumer Debt, especially its Unsecured forms of debt. From the Demographic Variables, Employment Status seems to hold some significance for separating debtors from non-debtors and from Economic Variables House Status appears to be an important predictor.

\begin{table}[htbp]
  \centering
  \caption{Important Variables in Random Forest Model}
    \begin{tabular}{lc}
    \toprule
    \multicolumn{2}{c}{Variable Importance} \\
    \midrule
    \textbf{Variable Importance} &  \textbf{Mean Decrease in Gini Index} \\    
    Impulsivity & 50.8253 \\
    Risk Aversion & 28.6954 \\
    Organisational Responsibility & 28.1296 \\
    Risk Management Belief & 27.8619 \\
    Planful Saving & 35.2052 \\
    Emp\_Status\_retired & 11.3329 \\
    House\_StatusOwn.outright & 24.1645 \\
    \bottomrule
    \end{tabular}%
  \label{tab 7}%
\end{table}%

Checking which attributes seem to have a measured variable importance above the mean we can also find among the important predictors additional attributes like Marital Status, Age/Sex, Social Grade, Education, Liquid Assets, Life Insurance and whether the households an Insurance at all. Most of these socio-economic attributes have been well researched in the literature \cite{wang2011demographics, kamleitner2012credit, stone2006indicators, kamleitner2007consumer} but some attributes like Liquid Assets and Life Insurance appear for the first time to be associated with Consumer Debt. It is of great surprise, however that Income and Household Income are not included in the important predictors. Income's ranking in variable importance is in the top 50\% but still below the average in this model, whereas throughout the literature it considered one of the most important predictors. Having in mind that the mean decrease in Gini Index might not be the appropriate method to identify the importance of each variable, we think that deeper and more careful analysis of the results is required in order to validate our findings from this model.

\section{Conclusion}
In this work, we explored the multifaceted nature of Consumer Indebtedness in a complete and detailed socio-economic dataset that contains Psychological information. After we identified and verified the Psychological Factors in the dataset we proceeded to the assessment of their contribution on modelling Consumer Indebtedness within a complete Data Mining Framework, with powerful classification models and reliable evaluation techniques.  It was shown that all the Psychological factors increase the performance of the models in a statistically significant sense in all cases. Especially Impulsivity, which is a well researched Psychological factor \cite{gathergood2012self,ottaviani2011impulsivity} seems to be the strongest predictor in our modelling. Our results further suggest that Consumer Debt Analysis has a lot to gain by embracing the studying of Psychological factors and by adopting Data Mining approaches in their methods and practices. Despite the fact that the transformations we performed on socio-economic variables were not able to make an impact in the modelling, they could prove a good dimensionality reduction method in the case of Demographics if that is needed in the analysis. Our final models achieved a good performance in separating debtors from non-debtors and a poor performance for predicting the level of debt.

Towards this direction, we would like to continue exploring the multifaceted nature of Consumer Indebtedness by examining the contribution of more factors like Financial Illiteracy which is mentioned in the literature \cite{gathergood2012self} to be associated with Consumer Debt. This way we can continue developing a more complete and accurate model for predicting the level of debt. Finally, a more elaborate and reliable method to identify the importance of individual variables needs to be researched developed in order to verify the findings of our work and get a better understanding of this complex model.


\section*{Acknowledgment}
We would like to thank John Gathergood, lecturer in
school of Economics of University of Nottingham for providing
us the DebtTracl survey.



%
\bibliography{Bibliography}{}
\bibliographystyle{plain}
\end{document}